\documentclass[conference]{IEEEtran}
\IEEEoverridecommandlockouts
\usepackage{cite}
\usepackage{amsmath,amssymb,amsfonts}
\usepackage{algorithmic}
\usepackage{graphicx}
\usepackage{textcomp}
\usepackage{xcolor}
\usepackage{times}
\usepackage{dblfloatfix}

\usepackage{algorithm}
\usepackage{algorithmic}
\usepackage{microtype}
\usepackage{graphicx}
\usepackage{subfigure}
\usepackage{booktabs} % for professional tables

\usepackage{amsmath}
\usepackage{amsthm}
\usepackage{amsfonts}
\usepackage{amssymb}
\usepackage{makecell}
\usepackage{bbm}

\usepackage{times}

\usepackage[textsize=footnotesize,color=green!40]{todonotes}

\usepackage{hyperref}

\newcommand{\R}{\mathbb{R}}

\newcommand{\states}{\mathcal{S}}
\newcommand{\actions}{\mathcal{A}}

\newcommand{\expect}[2]{\mathbb{E}_{#1}\left[ #2 \right]}

\def\BibTeX{{\rm B\kern-.05em{\sc i\kern-.025em b}\kern-.08em
    T\kern-.1667em\lower.7ex\hbox{E}\kern-.125emX}}
\begin{document}
\title{Membership Inference Attacks Against Temporally Correlated Data in Deep Reinforcement Learning}

\author{\IEEEauthorblockN{Maziar Gomrokchi\textsuperscript{\textsection}}
\IEEEauthorblockA{\textit{Department of Computer Science} \\
\textit{McGill University / Mila}\\
gomrokma@mila.quebec}
\and
\IEEEauthorblockN{Susan Amin\textsuperscript{\textsection}}
\IEEEauthorblockA{\textit{Department of Computer Science} \\
\textit{McGill University / Mila}\\
susan.amin@mail.mcgill.ca}
\and
\IEEEauthorblockN{Hossein Aboutalebi\textsuperscript{\textsection}}
\IEEEauthorblockA{\textit{Cheriton  School of Computer Science} \\
\textit{University of Waterloo / VIP Lab}\\
haboutal@uwaterloo.ca}
\and
\IEEEauthorblockN{Alexander Wong}
\IEEEauthorblockA{\textit{Department of Systems Design Engineering} \\
\textit{University of Waterloo / VIP Lab / DarwinAI}\\
a28wong@uwaterloo.ca}
\and
\IEEEauthorblockN{Doina Precup}
\IEEEauthorblockA{\textit{Department of Computer Science} \\
\textit{McGill University / Mila / DeepMind}\\
dprecup@cs.mcgill.ca}
}

\maketitle

\begingroup\renewcommand\thefootnote{\textsection}
\footnotetext{Equal contribution}
\endgroup

\thispagestyle{plain}
\pagestyle{plain}

\section{Abstract}

While significant research advances have been made in the field of deep reinforcement learning, there have been no concrete adversarial attack strategies in literature tailored for studying the vulnerability of deep reinforcement learning algorithms to membership inference attacks. In such attacking systems, the adversary targets the set of collected input data on which the deep reinforcement learning algorithm has been trained. To address this gap, we propose an adversarial attack framework designed for testing the vulnerability of a state-of-the-art deep reinforcement learning algorithm to a membership inference attack. In particular, we design a series of experiments to investigate the impact of temporal correlation, which naturally exists in reinforcement learning training data, on the probability of information leakage. Moreover, we compare the performance of \emph{collective} and \emph{individual} membership attacks against the deep reinforcement learning algorithm. Experimental results show that the proposed adversarial attack framework is surprisingly effective at inferring data with an accuracy exceeding $84\%$ in individual and $97\%$ in collective modes in three different continuous control Mujoco tasks, which raises serious privacy concerns in this regard.  Finally, we show that the learning state of the reinforcement learning algorithm influences the level of privacy breaches significantly.

\section{Introduction} \label{sec:intro}

Despite the recent advancements in the design and performance of deep reinforcement learning (deep RL) algorithms in complex domains (\cite{jumper2021highly,wurman2022outracing, silver2017mastering}), the vulnerability of these models to privacy breaches has only begun to be explored in the literature. In particular, while there have been a few studies on the vulnerability of deep RL models to adversarial attacks \cite{pan2019you, gleave2020adversarial, wu2021adversarial}, there has been no study on the potential membership leakage of the data directly employed in training deep RL models, which is known as membership inference attacks (MIAs). The potential success of such MIAs can have serious security ramifications in the deployment of models resulting from deep RL.

One of the major challenges in the implementation of MIAs in deep RL settings is the sequential and correlated nature of deep RL data points. Unlike in deep supervised settings, a data point in deep RL algorithms may consist of hundreds of correlated components in the form of tuples, all together forming a single trajectory. A successful MIA algorithm against a deep RL model should be able to learn not only the relation between the training and output trajectories but also the correlation between the tuples within each trajectory (data point). Another complication in this regard concerns the relationship between the training and prediction data points. In deep RL settings, batches of collected input data are used for training the deep RL policy. Thus, each output data point corresponds to every single data point in the training batches. This feature is in contrast with, for instance, that of data points in text generation problems (\emph{e.g.} machine translation or dialogue generation systems), where there is a direct (usually one-to-one) correspondence between the input and output sequential data points. Finally, RL algorithms are learning systems where the concept of labels is not defined as it is in supervised learning methods. Instead, during the learning phase, the deep RL agent receives reinforcement (aka rewards) from the environment as the outcome of the selected action. The deep RL agent uses the obtained rewards to learn the task and optimize its learning policy. These factors lead to complications in defining input-output pairs in training attack classifiers and subsequently establishing a meaningful relationship between the pair constituents.

Deep RL methods have a unique structural difference compared to deep supervised or unsupervised methods, \emph{i.e.} learning based on temporal correlation between the tuples in each trajectory and partial reinforcements the model receives upon interaction with the underlying environment. Even though deep RL models decorrelate input trajectories through an intermediate mechanism called \emph{replay buffer} (for more information, refer to the Background section), the inherent correlation between transition tuples still plays a significant role in the feature representation learned by the deep RL model \cite{mavrin2019deep}, hence the behaviour of the output policy. In this regard, two natural questions arise:

\begin{enumerate}
    \item How much information (concerning the training data points) can an adversary extract from the output of a trained deep RL model? 
    \item To what extent can an adversary benefit from feature correlation in the learned policy?
\end{enumerate}

This study presents the first black-box MIA against a deep RL agent to address these two questions. In our proposed adversarial attack framework, the target model is considered a black box; thus, the attacker does not have access to the internal structure of the deep RL agent. In particular, the attacker can only access the model output in the form of trajectories $\tau^{\mbox{\scriptsize out}}_T$ resulting from the trained policy $\pi_f$. We use \emph{batch off-policy reinforcement learning} setting, where the common practice is that an (unknown) \textit{exploration policy} (behaviour policy) $\pi_b$ collects private data points in the form of a batch of trajectories. The batch data is thereafter delivered to the deep RL algorithm in the form of independent trajectories (Markov chains) to train the target policy. In this setting, the RL agent decouples the data collection phase from the policy training phase (\emph{i.e.} off-policy). In the off-policy setting, the learning system is not tied to a particular exploration algorithm and ensures disjointedness between the training data sets provided for the RL algorithm in different settings. Off-policy setup is particularly preferred in designing MIA frameworks in black-box settings, where neither the internal structure of the target model nor the exploration policy used to collect the training trajectories is known to the adversary.

Our proposed attack framework tests the vulnerability of a state-of-the-art off-policy deep RL model to MIA in two modes: \emph{individual} and \emph{collective}. In the individual mode, the attacker's goal is to train a probabilistic model that infers the membership probability of a single trajectory $\tau^{\mbox{\scriptsize in}}_T$ given the trained policy $\pi_f$ and the initial state $s_0$. In this case, the goal is to measure the extent to which the adversary can exploit trajectory-level temporal correlation to reveal the presence of a trajectory in the training set. In the collective mode, the attacker's target is to predict the membership probability of a collection of data points. In this mode, the goal is to measure the extent to which the adversary is capable of exploiting not only the trajectory-level temporal correlation but also the batch-level correlation to reveal the presence of a trajectory in the training set. We show that the deep RL model is more vulnerable to collective MIA as in this mode, the attack classifier has access to more information. 

Moreover, we assess the vulnerability of the RL algorithm to MIA in terms of the learning state of the algorithm. Our results show that the cumulative amount of reinforcement the RL agent obtains in the course of training the policy is proportional to the level of its vulnerability to MIAs. Finally, to determine the role of data correlation in the vulnerability of the deep RL model to MIA, we disturb the correlation within the data points used to train the attack classifier and subsequently compare the impact of training the attacker with the resulting decorrelated trajectories on the performance of the attack classifier. We observe that the presence of correlation within the trajectories helps the adversary discern between the member and non-member data points with higher probability compared with the results obtained from the decorrelated case.

\section*{Background} \label{sec:background}
In this section, we provide the background information in two parts: \emph{i}) a general introduction to reinforcement learning systems, and \emph{ii}) membership inference attacks.

\subsection*{Reinforcement Learning}\label{sec:RL-intro}
In reinforcement learning (RL) systems, an agent learns a task through a sequence of trial and error and receives rewards through environmental interactions. The agent's task is formalized as a stochastic process that is described by a Markov Decision Process (MDP). An MDP is a tuple $\langle \states, \actions, \mathcal{P}, \mathcal{R},  p_0\rangle$ consisting of a set of states $\states$, a set of actions $\actions$, a transition probability kernel $\mathcal{P}: \states \times \actions \rightarrow \Pr(\states)$, a  reward function $\mathcal{R} : \states \times \actions \rightarrow R$, and an initial state distribution $p_0$ that characterizes the initial state of each episode. At each time step $t~=~0,1,2,\dots,T-1$, the agent is at the environment state $s_t \in \states$ and selects action $a_t \in \actions$ according to the policy $\pi(a_t|s_t)$. The policy $\pi: \states \rightarrow  \Pr(\actions)$ is the agent's action-selection strategy, which maps the current state to a distribution over actions and is updated throughout the learning process. Upon taking action $a_t$, the environment determines the agent's next state $s_{t+1}$ via the transition probability kernel $\mathcal{P}(s_{t+1}|s_t,a_t)$ and returns the reward $r_t$ computed by the reward function $\mathcal{R}(s_t,a_t)$. 

The RL agent's goal is to maximize the rewards received in the long run. The cumulative reward that the RL agent receives after time step $t$ is called \emph{return}, defined as $G^\pi_t := \sum_{k=0}^{\infty} \gamma^k r_{t+k+1}$, where the discount factor $\gamma \in [0,1]$ determines the weight of the future rewards. The value of each state at time $t$ under policy $\pi$ is called the \emph{state-value function} $V^\pi(s_t)$, and is defined as the expected return when the agent starts at $s_t$ and follows the policy $\pi$:
\begin{align}
    V^\pi(s_t) = \mathbb{E}_\pi\{G_t|s_t\}.\label{eq:state_value}
\end{align}
Similarly, we can determine the value of a state $s_t$ and action $a_t$ taken at time $t$ (on the condition that we follow the policy $\pi$ afterwards) using the notion of \emph{action-value function} $Q^\pi(s_t,a_t)$, defined as
\begin{align}
    Q^\pi(s_t,a_t) = \mathbb{E}_\pi\{G_t|s_t,a_t\}.\label{eq:action_value}
\end{align}
The ultimate goal of the RL agent is to learn an effective policy that maximizes the value functions.

In the context of RL, a data point in a batch of data is a sequence of temporally correlated tuples $(s_t,a_t,r_t,s_{t+1})$ that denote the history of the RL agent's interaction with the environment. This sequence of tuples is often referred to as \textit{trajectory}, and is denoted as

\begin{align}
   \tau_T=&(s_0,a_0,r_1,s_1), (s_1,a_1,r_2, s_2),\dots\\ \nonumber
   &,(s_{T-2},a_{T-2},r_{T-1},s_{T-1}).
\end{align}

\begin{figure*}[htbp]
\begin{center}
\includegraphics[width=0.8\textwidth]{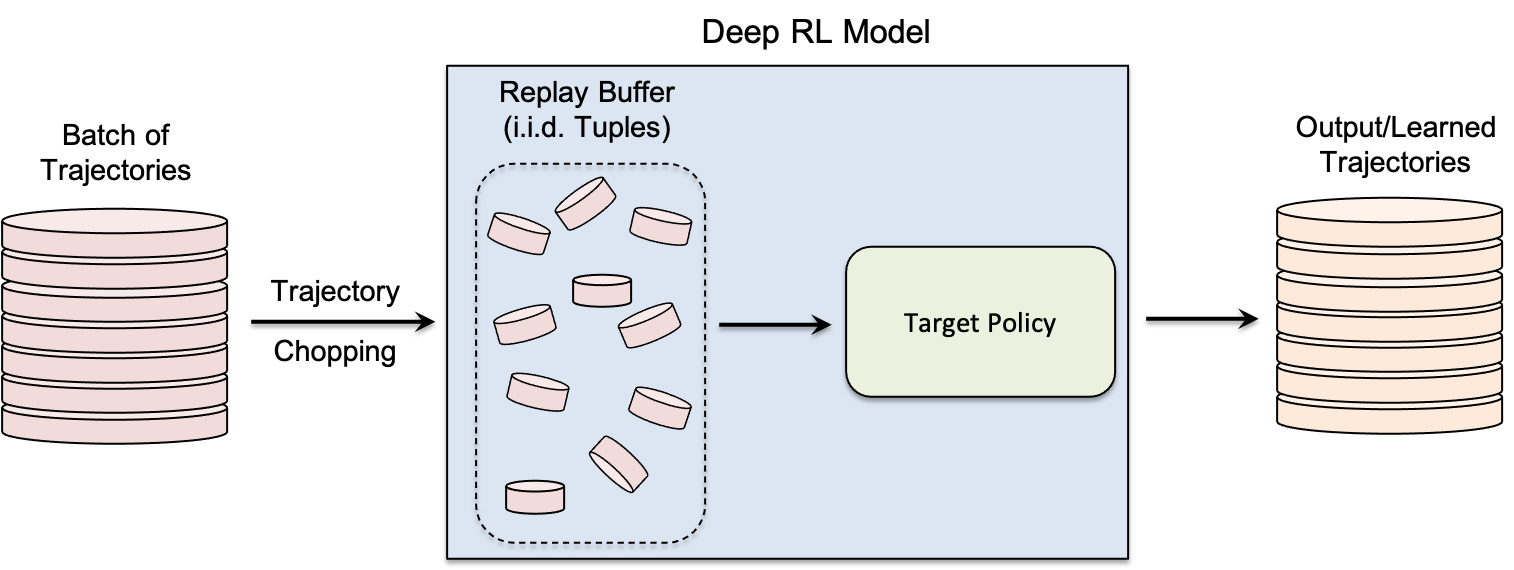}
\end{center}
\caption{Batch off-policy RL learning architecture. An external behaviour policy generates a batch of trajectories composed of transition tuples (state, action, reward, new state). The trajectories are subsequently passed to the deep RL model. The replay buffer mechanism, as an internal part of the model, decorrelates each trajectory into a collection of i.i.d. transition tuples and then uses them in the form of mini-batches to train the target policy. 
}\label{fig:rl-arch}
\end{figure*}

RL agents do not have knowledge of the environment at the very initial stage of learning and acquire the necessary experience through continued interactions with the environment. An RL agent can acquire the necessary information in two ways: \textit{on-policy} and \textit{off-policy}. In the on-policy setting, the agent uses the target policy that is trained so far to obtain data through interaction with the environment. In the off-policy setting, on the other hand, the agent receives an input batch of data in the form of trajectories provided by an exploratory agent (\emph{i.e.} \textit{behaviour policy} $\pi_b$), and subsequently uses the acquired data to train the \emph{target policy} $\pi_f$ (exploitation). The output of the trained target policy consists of data points (trajectories) produced as a result of the interaction between the target policy and the environment (Figure \ref{fig:rl-arch}). From the privacy point of view, since the private data is assumed to exist a priori, off-policy methods are natural choices to be analyzed in this regard. Figure \ref{fig:rl-arch} presents a schematic of off-policy deep RL architecture.

The fact that the input trajectories in off-policy deep RL models are temporally correlated necessitates the use of a mechanism that converts the input data to \emph{i.i.d.} samples before passing it to the deep network. A widespread and fundamental data management mechanism that has become an inevitable part of the existing off-policy deep RL models is \textit{experience replay buffer} or \textit{replay buffer}. The main intuition behind the application of replay buffer in deep RL lies at the heart of RL theory. It is well-studied that fundamental RL algorithms (\emph{e.g.} Q-learning) easily diverge in the case of linear function approximation \cite{sutton1998reinforcement, fairbank2012divergence}. The solution that replay buffer offers to the divergence problem of these algorithms is to decorrelate the input trajectories and subsequently treat each transition tuple as an i.i.d. sample point (Figure \ref{fig:rl-arch}). This intermediate decorrelation step significantly improves data efficiency and helps the deep RL algorithm converge to the optimal policy according to the law of large numbers. Moreover, it allows the deep RL algorithm to benefit from mini-batch training and shuffling techniques, which are proven to improve the performance of deep RL algorithms significantly \cite{mnih2015human,zhang2017deeper,silver2017mastering,liu2018effects,fedus2020revisiting}.

\subsection*{Membership Inference Attack}\label{sec:intro-MIA}

In machine learning, a membership inference attack (MIA) or tracing attack \cite{dwork2017exposed,shokri2017membership} is a form of adversarial attack that is designed to infer the presence of a particular data point in the training set of a target model. The central intuition in the design of MIAs is that publicly available trained models tend to exhibit higher confidence in their predictions of the individuals who participated in the training data. Consequently, the members of training sets are vulnerable to privacy threats \cite{shokri2017membership, salem2018ml}. The main challenge for the adversary in MIAs is to design a classifier compatible with the target model domain setting and decide whether a particular data point was part of the training set given the training target model's output. Attackers employ different MIA design strategies based on: i) the adversary's knowledge level of the parameters in the target model (Label-only strategy) and ii) the adversary's knowledge level of the training data (Shadow model technique). 

In the \textit{Label-only} strategy \cite{yeom2018privacy,choquette2021label}, the attacker only relies on model predictions and discards the model's confidence scores. In this technique, the attacker uses the generalization gap (the difference between the train and test accuracy) in the attack model as the main driver in inferring the membership of individuals used in training the target model. The label-only technique was first introduced by Yeom et al. \cite{yeom2018privacy} and was subsequently extended by Choquette \emph{et al.} \cite{choquette2021label} to show how the label-only technique can improve the existing attack baselines. As the notion of the label is not defined in the general RL setting, the label-only technique cannot be applied here in devising MIAs against RL models.

\textit{Shadow model} technique \cite{shokri2017membership} is known as an effective and practical approach for designing MIA models. Shadow models are parallel local models trained on data sets often sampled from the same distribution as the underlying distribution of the private data. In this method, the adversary trains the shadow models with complete knowledge of the training set. Thus, using the auxiliary membership information and the trained shadow models, the adversary can build a membership classifier that identifies whether an individual has participated in the training of similarly trained models. 

In the training phase in both \emph{label-only} and \emph{shadow model} techniques, the adversary should have access to the model output labels and the training data true labels. However, the sequential nature of the training and output data points and the temporal nature of model training make the design of MIAs for RL models fundamentally different. Moreover, the presence of replay buffer as an inevitable part of off-policy deep RL models adds another level of complexity to the design of MIAs, as this intermediate transformation phase adds a new source of noise to the data from the attacker's perspective.  
\section*{Related Work} \label{sec:related}

MIAs were used for the first time against machine learning systems by Shokri \emph{et al.} \cite{shokri2017membership}. In the following years, extensive studies were performed on the application of MIAs against supervised (\cite{shokri2017membership,9230385,yeom2018privacy, salem2018ml,song2021systematic}) and unsupervised (\cite{HayesMelisDanezis,hilprecht2019monte, chen2020gan}) machine learning models, surveyed comprehensively by Hu \emph{et al.} \cite{hu2021membership}, and Rigaki and Garcia \cite{ rigaki2020survey}. This section reviews the existing attack models against supervised and unsupervised models trained on sequential data.

MIAs have been executed against aggregate location time-series \cite{pyrgelis2017does, PyrgelisTC18, pyrgelis2020measuring}. For the first time, Pyrgelis \emph{et al.} \cite{pyrgelis2020measuring} studied the impact of different spatial-temporal factors that contribute to the vulnerability of time-series-based algorithms to MIAs. MIAs have also been studied in the context of text generation problems \cite{song2019auditing, hisamoto2020membership}, where the attacker's goal is to identify whether or not a specific sequence-to-sequence or sequence-to-word pair is part of the input training data of a machine translation engine, a dialogue system or a sentimental recommendation system. The structure of machine learning algorithms with sequential data differs from that of classic classification tasks in the input and prediction types. While inputs and outputs in standard classification problems have fixed sizes, they are chains of correlated elements with variable lengths in sequence generation tasks. This difference poses a fundamentally different approach to designing MIAs against sequence generation tasks. The knowledge of output space distribution is no longer valid for the attack classifier since the output length may vary from one model to another. To tackle this challenge, Song and Shmatikov \cite{song2019auditing} assume access to a probability distribution over output-space vocabularies. They \cite{song2019auditing} split their proposed attack model into two phases, shadow model training and audit model training. In the shadow model training phase, the attacker trains multiple shadow models assuming that the attacker has access to a generative model that generates a sequence of vocabularies. In the audit training phase, the attacker uses the rank of the words produced by the target model instead of the output probability distribution. The central assumption is that the gap observed between the trained model rank predictions depends on word frequencies in the training and test sequences. In a similar study,  Hisamoto \emph{et al.} \cite{hisamoto2020membership} address MIA against sequence-to-sequence models in a setting where the adversary is agnostic to the word sequence distribution. In their work, the attacker is equipped with a generative model for different translation subcorpora, an alternative for output word sequence distribution. 

While in models trained on sequential data, the input-output relation is well defined and deterministic, in deep RL models, the output data are generated through the trained policy; thus, each output sequence can be considered as evidence for the entire input dataset. Therefore, one requires a fundamentally different approach in designing MIAs against RL algorithms. To the best of our knowledge, there is no prior work in the context of deep RL that addresses the problem of membership inference at a microscopic level, where the attacker infers the membership of a particular data point in the training set of deep RL models \cite{hu2021membership, rigaki2020survey}.
\section*{Methods and Experiments} \label{sec:attack}

In our proposed adversarial attack framework, we successfully conduct MIA against deep RL in a black-box setting, where only the model output is accessible to external users. The deep RL model interacts with an environment whose distribution of initial states, state space $\states$ and action space $\actions$ are common knowledge, an assumption widely accepted in the RL community \cite{sutton1984temporal,vietri2020private,szepesvari2010algorithms}. In this section, we first explain the general setting of the problem and subsequently introduce our attack platform and our proposed method of data formatting for training the attack models. We further mention the different settings we have considered in our experimental design. Finally, we provide our choices of performance measures to assess the behaviour of the attack model. 

\subsection*{General Setup}
\begin{figure*}[htbp]
\begin{center} 
\includegraphics[width=0.7\textwidth]{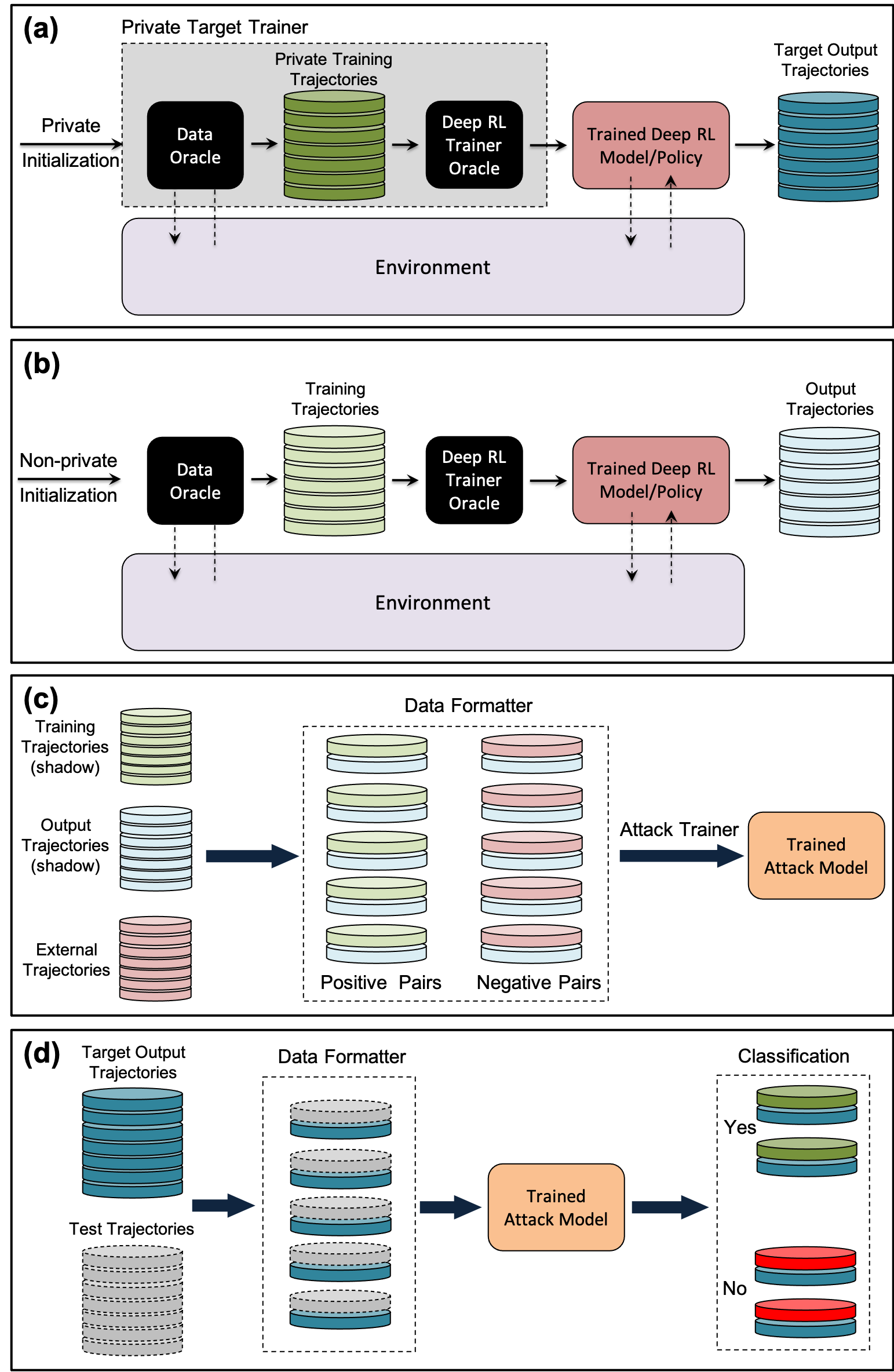} 
\end{center} 
\caption{The proposed black-box MIA architecture. \textbf{(a)} Private deep RL model training: the black-box exploration engine (data oracle) interacts with the environment and provides private training trajectories for the black-box deep RL model trainer. The trained deep RL model is subsequently used to output target trajectories via interaction with the environment. \textbf{(b)} Shadow training: the data oracle is used to produce non-private training trajectories to be used as input to train the deep RL trainer oracle. The output trajectories are subsequently generated by the trained deep RL model. \textbf{(c)} Training the attack classifier: the input and output trajectories obtained in part (b) are paired together in \emph{data formatter} to provide \emph{positive} training pairs. Another set of trajectories, which has not been used in training the shadow model, is used with the output trajectories from part (b) to create \emph{negative} training pairs. The attack model is subsequently trained using the paired trajectories with the corresponding \emph{positive} and \emph{negative} labels. \textbf{(d)} Membership inference attack: the target output trajectories from part (a) are paired with sample test trajectories in the data formatter. The trained attack model subsequently uses the pairs to infer the test set trajectories that were used to train the private deep RL model.} \label{fig:attack-arch}
\end{figure*}

We propose an adversarial attack method for studying the vulnerability of the deep RL algorithm to MIA in a black-box setting, where the attacker's access to the model is limited to the output trajectories of the model trained on a private batch of input trajectories. Figure~\ref{fig:attack-arch} depicts the general framework of our proposed black-box attack against deep RL algorithms.

The two important oracles that should accompany the end-to-end design of a black-box attack model in off-policy deep RL are i) the data oracle $\mathcal{O}_{\mbox{\scriptsize data}}$, and ii) the model trainer oracle $\mathcal{O}_{\mbox{\scriptsize train}}$. The data oracle interacts with the environment and returns a set of independent and identically distributed (i.i.d.) training trajectories (Markov chains) for the model trainer oracle $\mathcal{O}_{\mbox{\scriptsize train}} $ (see Figures \ref{fig:attack-arch}~(a, b)). The data oracle is a black box which is equipped with a set of unknown exploration policies. To train the target model, whose training input is of the adversary's interest, the data oracle is initialized privately (see Figure \ref{fig:attack-arch}~(a)), leading to the generation of a batch of private training data points in the form of trajectories. The model trainer oracle is agnostic to the exploration policy used for the data collection. The training data batch is passed to the deep RL trainer oracle, and the resulting trained model is made publicly available for data query. Our experimental framework can adopt any of the existing off-policy batches deep RL models as the deep RL trainer oracle. In this study, we choose to work with the state-of-the-art Batch-Constrained deep Q-learning (BCQ) \cite{fujimoto2019off} model, which is widely used as the basis of other deep RL algorithms and exhibits remarkable performance in complex control tasks. Structurally, BCQ trains a generative model on the input trajectories such that the model learns the relationship between the visited states in the input trajectories and the corresponding actions taken. The BCQ algorithm subsequently uses the developed generative model to train a deep Q-network, which ultimately learns to sample the highest-valued actions similar to the ones in the input trajectories. 

We use the \emph{shadow model} \cite{shokri2017membership} training technique to acquire the data needed for training the attack classifier. In this method, through the data oracle, the attacker provides the deep RL trainer oracle with a set of \textit{non-private }training trajectories (Figure \ref{fig:attack-arch}~(b)), on which the deep RL model is trained. The attacker subsequently queries output trajectories from the trained deep RL model and passes the training and output trajectories to the data formatter (Figure \ref{fig:attack-arch}~(c)). In this step, the training-output trajectories are augmented into pairs and are subsequently labelled as $1$ in positive pairs and $0$ in negative ones depending on whether or not the trajectories belong to the same trained model. Finally, the \textit{attack trainer} trains a probabilistic classifier that takes as input the pairs of trajectories prepared by the data formatter and returns a trained probabilistic attack classifier that is subsequently used to infer the membership of target input trajectories (Figure \ref{fig:attack-arch}~(d)).

Since the attack training data collected by the data oracle $\mathcal{O}_{\mbox{\scriptsize data}}$ and prepared by the data formatter is of a sequential nature, we need to adopt an attack model that is compatible with time-series data. The classifier should minimize the expected loss, defined as
\begin{align} \label{eq:classification-loss}
    \expect{\mathcal{D}}{l(A_{D,\theta}(., \pi_f), g(.))} \approx \frac{1}{|D|} \sum_{\tau \in D} l(A_{\boldsymbol{\theta}}(\tau, \pi_f), g(\tau, \pi_f)),
\end{align}
\noindent where $g(.)$ is the function that assigns labels to the formatted pairs, $A(.)$ is the parameterized classifier, and $l(.)$ is the loss function adopted by $A$. The dataset $D$ contains a set of i.i.d. trajectories drawn from $\mathcal{D}$, and $\pi_f$ denotes the policy trained on $D$. The goal of the attacker is to train a classifier that learns a parameter vector (or network) $\theta^*$ that minimizes the loss function. The following sections provide more details regarding the data formatter and attack classifier.

\subsection*{Experimental Setup}\label{sec:setup}
In our experimental design, we study the vulnerability of the deep RL model to MIAs in terms of the following factors:

1) \emph{the membership inference mode (collective vs. individual MIA) -} In the individual mode, the adversary's goal is to infer the membership of \emph{single} training data points (trajectories), while in the collective mode, the adversary's target is a \emph{batch} of trajectories used in the training of the deep RL model. In this experimental setup, we aim to address two scenarios in which a participant's identity is revealed. In the individual mode, a trajectory reveals a user's identity, while in the collective mode, a collection of trajectories represents the user's identity.

2) \emph{the maximum trajectory length $T_{\mbox{\scriptsize max}}$ within each episode -} The value of $T_{\mbox{\scriptsize max}}$ is determined and fixed by the environment during data collection and model training. In particular, the RL agent's trajectory in each episode ends when either the agent arrives at an absorbing state at $T<T_{\mbox{\scriptsize max}}$ or the number of time steps $T=T_{\mbox{\scriptsize max}}$. Larger $T_{\mbox{\scriptsize max}}$ corresponds to larger values of return (cumulative reward), thus an improved deep RL policy. 

3) \emph{the level of correlation within the input trajectories used to train the attack classifier -} In the case of individual MIA, we study the performance of our proposed attack classifier in two modes: 1) \emph{correlated} mode, where the adversary is trained on pairs with undisturbed input trajectories, 2) \emph{decorrelated} mode, where in the attack training phase, the input trajectory is formed by sampling tuples at random from the whole batch. This set of experiments provides useful information regarding the effect of the correlation level within the input trajectories on the performance of the attack model. 

Below is a detailed description of the environments used in our experimental design, the data formatting technique, and the attack architecture.

\textbf{Environments and RL Setting-} We assess the algorithm on OpenAI Gym environments \cite{brockman2016openai} powered by MuJoCo physics engine \cite{todorov2012mujoco}, which are standard tasks adopted by many recent RL studies \cite{lillicrap2015continuous, haarnoja2018soft, fujimoto2018addressing, henderson2018deep, franccois2018introduction}. Gym provides a variety of simulated locomotion tasks with different action and state space dimensionalities. Here, we train the deep RL agent on three high-dimensional continuous control tasks: \textit{Hopper-v2} ($\actions\subset\mathbb{R}^3$ and  $\states\subset\mathbb{R}^{11}$), \textit{Half Cheetah-v2} ($\actions\subset\mathbb{R}^6$ and $\states\subset\mathbb{R}^{17}$), and \textit{Ant-v2} ($\actions\subset\mathbb{R}^8$ and $\states\subset\mathbb{R}^{111}$). Starting from virtually zero knowledge of how each task works, the deep RL model's goal is to teach the Hopper how to hop, the HalfCheetah how to run, and the Ant how to walk as fast as possible. We use the Deep Deterministic Policy Gradient (DDPG) algorithm \cite{lillicrap2015continuous} as the data oracle $\mathcal{O}_{\mbox{\scriptsize data}}$ and Batch-Constrained Deep Q-Learning (BCQ) \cite{fujimoto2019off} as the batch off-policy deep RL method used in the trainer oracle $\mathcal{O}_{\mbox{\scriptsize train}}$. 

\begin{figure*}[htbp]
\begin{center} 
\includegraphics[width=0.9\textwidth]{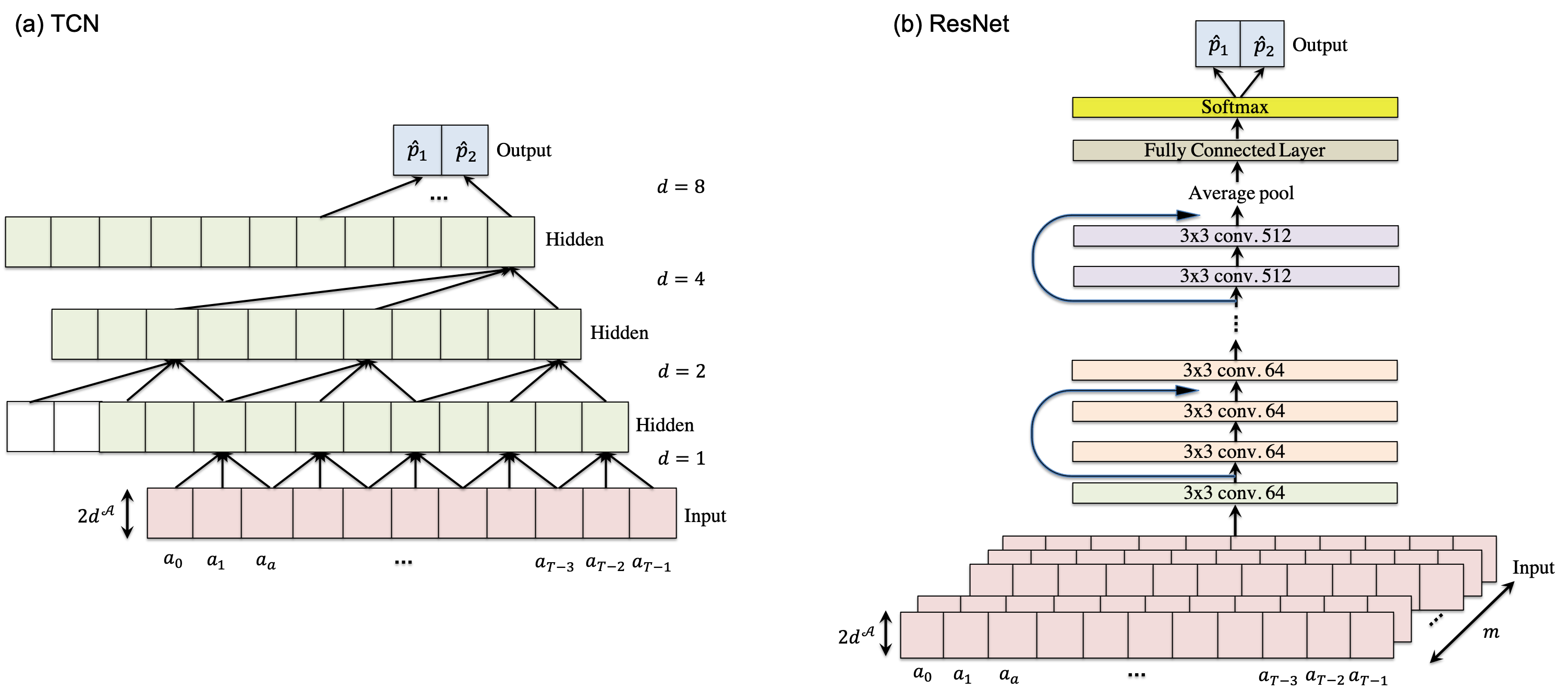} 
\end{center} 
\caption{The network architecture of TCN \textbf{(a)} and ResNet \textbf{(b)} used in the individual and collective MIAs, respectively.} \label{fig:network-arch}
\end{figure*}

\textbf{Data Augmentation- } Each trajectory starts with an initial state $s_0$ drawn from the available distribution of initial states in the environment, followed by action $a_0$ selected and taken by the RL agent. The environment subsequently takes the agent to the next state $s_1$ and returns the reward $r_1$. The agent's next choice of action is based on $s_1$, and this cycle continues until the trajectory ends at $s_T$. In other words, the initial state $s_0$ plays a significant role in determining the sequence of actions taken by the RL policy and the consequent states and rewards. Thus, to prepare training pairs to train the attack classifier, we pair the training and output trajectories that have the same initial states, fixing the starting point of the two trajectories in a pair. Moreover, as the RL agent interacts with MDP, the resulting trajectory is a Markov chain, \emph{i.e.} every state and reward in the trajectory is the direct consequence of the previous state and action. Therefore, we choose to remove states and rewards from the trajectories, keep the actions in the trajectory, and use them in the pairing process.

Each task is equipped with a set of absorbing states $\mathcal{B} \in \states$. An absorbing state is a state that leads to the termination of an agent's chain of interactions with an environment. Due to the presence of absorbing states in the environment, the generated trajectories have different lengths. To pair the training and output action trajectories obtained from the deep RL model, we need to either increase the length of shorter action trajectories to match that of the longest one or clip longer action trajectories to a pre-determined length. Based on the desired length, we choose to repeat the last action in shorter action trajectories for the required number of times and trim longer trajectories. Each action trajectory is a $d^{\actions}\times T$ dimensional array, where $d^{\actions}$ is the dimension of action space, and $T$ is the total number of actions in the trajectory. Every output action trajectory is concatenated with a training trajectory such that the resulting pair is a $2d^{\actions}\times T$ dimensional array. The pairs are subsequently passed to the attack classifiers in multi-dimensional arrays $\R^{2d^{\actions} \times T }$ and $\R^{2d^{\actions} \times T \times m }$ in individual and collective modes, respectively. The value $m$ refers to the number of pairs in each batch in the collective mode.

\textbf{Attack Classifier Architecture- } We use Temporal Convolutional Networks (TCNs) \cite{bai2018empirical} as the classifier for individual MIA, and Residual Network (ResNet) \cite{he2016deep} deep architecture for collective MIA. Figure \ref{fig:network-arch} shows a schematic of TCN (Figure \ref{fig:network-arch}(a)) and ResNet (Figure \ref{fig:network-arch}(b)) architectures.

\emph{Individual-Mode Attack Classifier Architecture- } As both training and output trajectories of RL models are composed of temporally correlated transition tuples, the choice of attack classifier must utilize the input-level temporal correlation in its feature representation. TCNs are structurally designed to utilize the inherent temporal correlation in the training data through a hierarchy of temporal convolutions architecture. In this regard, TCN employs a 1D fully-convolutional network (FCN) architecture \cite{long2015fully}, where each of its hidden layers has the same length as the input layer (Figure \ref{fig:network-arch}(a)). The main advantage of TCN is its ability to use dilation in convolution layers to keep the long-range temporal dependency and increase the receptive field of the convolutional layers. In the individual MIA mode, since the input data to the classifier is a pair of temporally correlated tensors (i.e. $\R^{2d^{\actions} \times T }$), the long-range correlation between input tuples within each trajectory is well-aligned with the input structure of TCNs. For more information on the internal structure of TCN architecture, refer to the Appendix.

\emph{Collective-Mode Attack Classifier Architecture- }
 In this case, while more information is accessible to the attacker, it requires a more complex learning architecture and more sophisticated hyper-parameter tuning to exploit the cross-correlation among the training trajectories and the temporal correlation within a trajectory. In the collective mode, our input is in the form of three-dimensional tensor (\emph{e.g.} $\R^{2d^{\actions} \times T \times m }$). Unlike the individual MIA mode, which involves 2-dimensional inputs, in the collective MIA mode, we have another dimension $m$ for the number of trajectories in each batch of trajectories, similar to the data structure used in image classification problems \cite{he2016deep, simonyan2014very, huang2017densely}. Thus, we use the deep residual network (ResNet) architecture  \cite{he2016deep} because of its inherent compatibility with data sets with temporally deep structures. ResNets are popular for solving standard computer vision problems \cite{he2016deep, minaee2021image, zhao2019object}. 
\begin{figure*}[htbp]
\begin{center}
\includegraphics[width=0.8\textwidth]{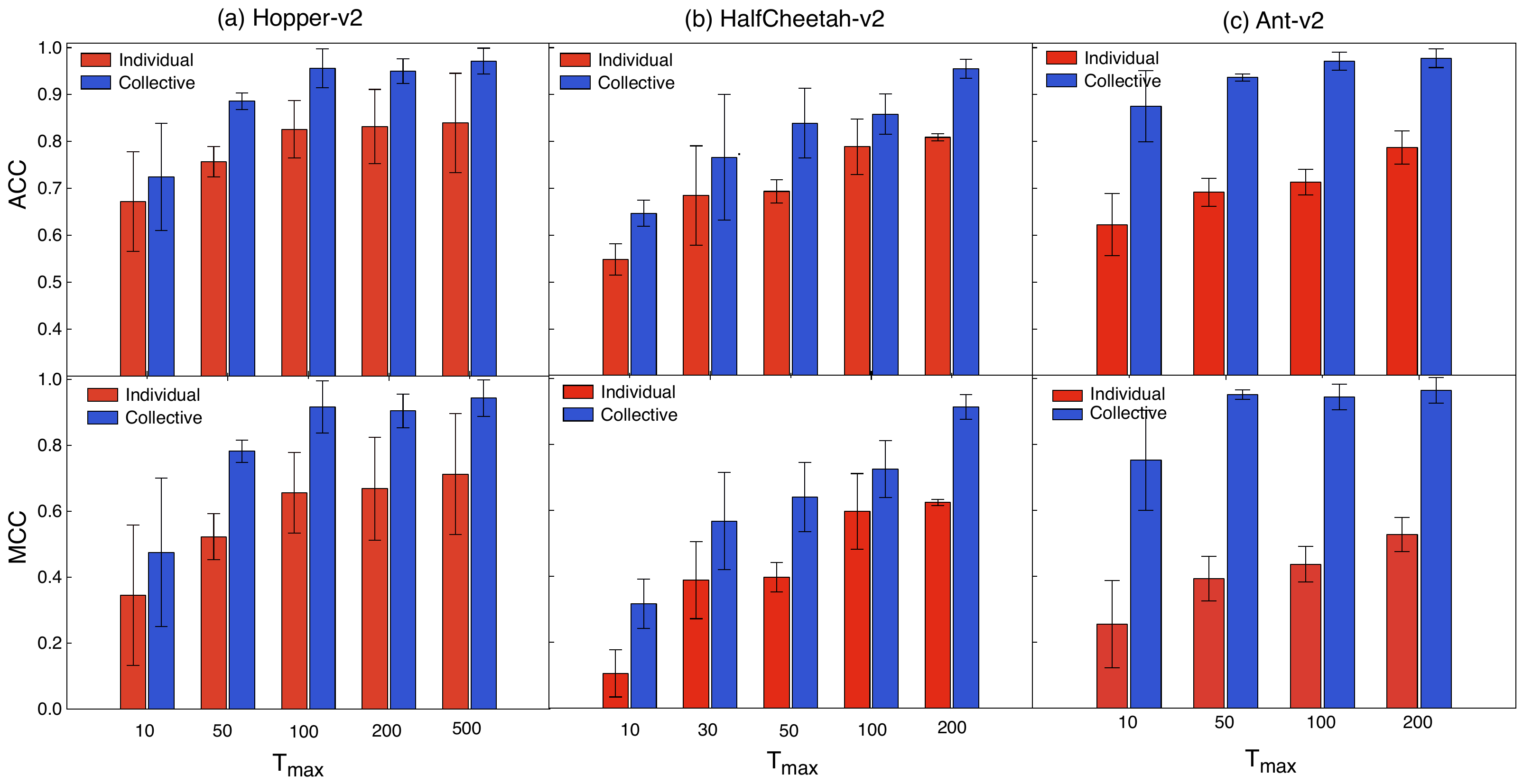} 
\end{center}
\caption{The performance of the attack classifiers in tasks Hopper-v2 \textbf{(a)}, HalfCheetah-v2 \textbf{(b)}, and Ant-v2 \textbf{(c)} in individual and collective attack modes. Each data point is determined from the average result of $5$ separate runs. The error bars depict the error on the mean for ACC (\emph{top}) and MCC (\emph{bottom}) in the corresponding runs. The batch size $m = 50$ in the collective mode.}\label{fig:metrics}
\end{figure*}
\subsection*{Performance Metrics}

We adopt the standard performance metrics used in the classification literature \cite{sokolova2009systematic} to evaluate the performance of our proposed attack models. We measure the performance of the attack classifier with the following metrics:

\textbf{\emph{{Overall accuracy}} (ACC)}, which captures the overall performance of the attack classifier and is calculated as follows,
\begin{align}
    \mbox{ACC} = \frac{\mbox{TP} + \mbox{TN}}{\mbox{TP} + \mbox{TN} + \mbox{FP} + \mbox{FN}},
\end{align}
where TP (true positives) denotes the number of correctly recognized positives, and TN (true negatives) shows the number of correctly recognized negative ones. The two other quantities, false positives FP and false negatives FN indicate the number of incorrectly recognized positives and negatives, respectively.

\textbf{\emph{Precision} (PR)}, which shows the fraction of pairs classified as matching pairs that are indeed coming from the same model, and is written as, $\mbox{PR} = {\mbox{TP}}/{\mbox{TP} + \mbox{FP}}$.

\textbf{\emph{Recall} (RE)}, which measures the fraction of matching pairs that the attack classifier can infer correctly, and is computed as, $\mbox{RE} = {\mbox{TP}}/{\mbox{TP} + \mbox{FN}}$.

\sloppy\textbf{\emph{F1 score} (F1)}, which is the harmonic mean of the precision (PR) and recall (RE), and is calculated as~
$\mbox{F1} = ({2.\text{\mbox{PR}}.\text{\mbox{RE}}})/({\text{\mbox{PR}} + \text{\mbox{RE}}})$.

\textbf{\emph{Matthews Correlation Coefficient} (MCC)} \cite{matthews1975comparison}, which calculates the correlation between the predicted and the true classification labels, and is defined as,
\begin{align}
\mbox{MCC} = \frac{\mbox{TP}.\mbox{TN} - \mbox{FP}.\mbox{FN}}{\sqrt{(\mbox{TP}+\mbox{FP})(\mbox{TP}+\mbox{FN})(\mbox{TN}+\mbox{FP})(\mbox{TN}+\mbox{FN})}}.
\end{align}

MCC is an effective and meaningful combination of all four quantities TP, TN, FP, and FN, and ranges from $-1$ to $1$. The closer MCC is to $1$, the better the model performs  \cite{chicco2021matthews}. MCC$=0$ shows that the model is a random guesser. The other evaluation metrics ACC, PR, RE, and F1 vary in the $[0, 1]$ range. In a well-performing model, all of these evaluation metrics have values close to $1$. Finally, to show the performance of our proposed MIA classifiers in individual and collective modes at different classification thresholds $\theta$, we plot receiver operating characteristic (ROC) curve, which shows the changes of recall RE as a function of \textit{False Positive Rate} $\mbox{FPR} = \mbox{FP}/(\mbox{FP} + \mbox{TN})$ for different values of $\theta$.

\section*{Results and Discussion}

This section presents and discusses the results of different experimental scenarios to capture the interdependence between different parameters that affect the accuracy of membership inference in deep RL settings.    

\begin{figure*}[h!]
\begin{center}
\includegraphics[width=0.8\textwidth]{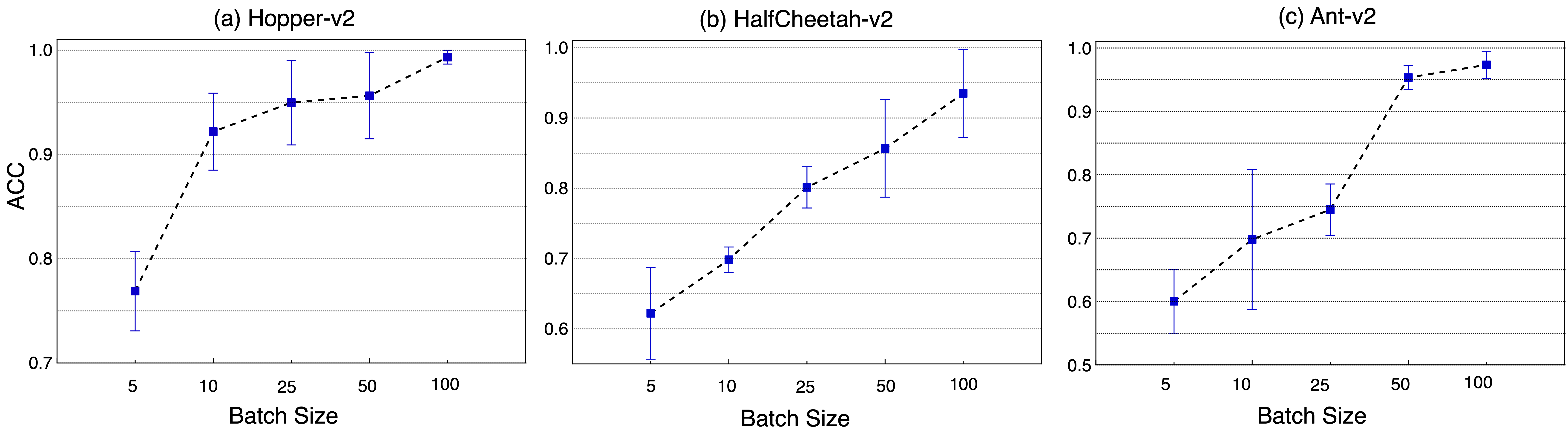} 
\end{center}
\caption{The MIA accuracy in Hopper-v2 \textbf{(a)}, HalfCheetah-v2 \textbf{(b)}, and Ant-v2 \textbf{(c)} in the collective attack mode for different batch sizes. Each data point is determined from the average result of $5$ separate runs. The error bars depict the error on the mean. The maximum trajectory length $T_{\mbox{\scriptsize max}} = 100$.}\label{fig:buffer_size}
\end{figure*}
\begin{figure*}[!b]
\begin{center}
\includegraphics[width=0.9\textwidth]{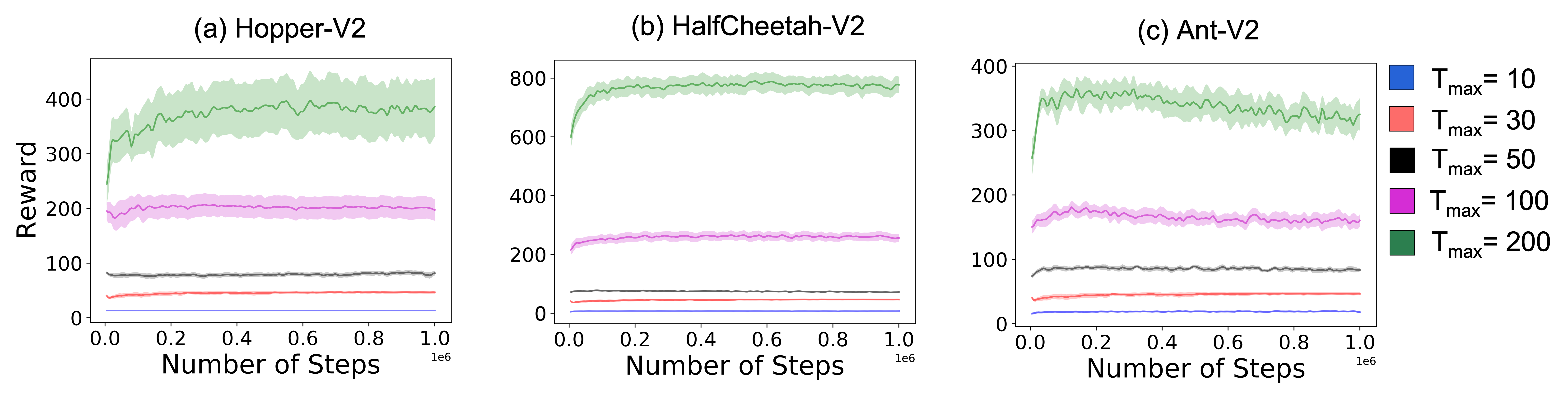} 
\end{center}
\caption{Deep RL curves in three high-dimensional locomotion tasks Hopper-v2 \textbf{(a)}, HalfCheetah-v2 \textbf{(b)}, and Ant-v2 \textbf{(c)}. The graphs depict the performance of the deep RL model as a function of time for different maximum trajectory lengths $T_{\mbox{\scriptsize max}}$. The plots are averaged over $5$ random seeds. The performance of the deep RL policy is assessed every $5000$ step over $1000000$ time steps.}
\label{fig:mujoco-tasks}
\end{figure*}
\subsection*{Collective vs. Individual MIAs} \label{sec:collective-individual}
Using different classification metrics, we assess the behaviour of TCN and ResNet attack classifiers in predicting the membership probability of individual and collective data points, respectively. Figure \ref{fig:metrics} presents the performance of the classifiers TCN and ResNet in Hopper-v2 (Figure \ref{fig:metrics}(a)), HalfCheetah-v2 (Figure \ref{fig:metrics}(b)), and Ant-v2 (Figure \ref{fig:metrics}(c)) in terms of ACC and MCC for different maximum trajectory lengths $T_{\mbox{\scriptsize max}}$. The full report of their performance in these three tasks is provided in Tables 1-3 in the Appendix (Section \ref{sec:tbl}). The results show that our proposed attack framework can infer the RL model training data points with high accuracy (\emph{e.g.} $>0.8$ in the individual and $>0.9$ in the collective mode for $T_{\mbox{\scriptsize max}} \geq 100$ in Hopper-v2), indicating a high risk of privacy invasion. Moreover, the results reveal that for a fixed $T_{\mbox{\scriptsize max}}$, the adversary infers collective data points with significantly higher accuracy than the accuracy value in the individual mode. For example, in the Hopper-v2 task, the membership inference accuracies in the collective mode for $T_{\max}$ are more than $12\%$ higher than those in the individual mode. This observation shows that the deep RL algorithm is more vulnerable to MIA in the collective mode, which is expected since more information is provided to the attack classifier through a batch of data points instead of one. In particular, in the collective mode, the adversary can capture the collective properties of the training data points and their relationship with the output trajectories, which could be veiled in one individual trajectory. 

To further study the effect of \emph{batch size} on the performance of MIA in the collective mode, we conduct MIAs against the deep RL agent for different batch sizes (Figure \ref{fig:buffer_size}). A closer analysis of the two figures (Figure \ref{fig:metrics} and Figure \ref{fig:buffer_size}) reveals that while larger batch sizes correspond to higher level of deep RL training members' vulnerability to MIA, batch sizes $m\leq 5$ in Hopper-v2, $m< 25$ in HalfCheetah-v2, and $m\leq 10$ in Ant-v2 lead to smaller values of inference accuracy compared with those in the individual mode. We believe that this difference in the performance of the adversary between the two cases corresponds to the different structures used in the individual and collective modes (\emph{i.e.} TCN and ResNet). In particular, our results show that the effectiveness of the ResNet classifier in inferring the membership of the data points surpasses that of TCN in larger batch sizes.

\subsection*{The Impact of $T_{\mbox{\scriptsize max}}$}
We test the performance of attack classifiers against the target model for different values of $T_{\mbox{\scriptsize max}}$ in a set of experiments. As the environment is unvarying, the value of $T_{\mbox{\scriptsize max}}$ remains unchanged throughout each experiment. Our observations presented in Figure \ref{fig:metrics} show that as $T_{\mbox{\scriptsize max}}$ increases, the accuracy ACC of the attack classifiers in inferring the training data points in both individual and collective modes improves. Moreover, our results show consistent improvement of MCC as a function of $T_{\max}$ in all three environments Hopper-v2, Half Cheetah-v2, and Ant-v2, which is consistent with the changes in ACC. Note that as MCC utilizes all four values in the confusion matrix, it provides a more reliable and robust measure compared to the other metrics (for additional results and comparison, refer to the tables provided in the Appendix).

Maximum trajectory length $T_{\mbox{\scriptsize max}}$ plays a significant role in the performance of deep RL models. Figure \ref{fig:mujoco-tasks} illustrates the learning curves for the deep RL model in Hopper-v2 (Figure \ref{fig:mujoco-tasks}(a)), HalfCheetah-v2 (Figure \ref{fig:mujoco-tasks}(b)), and Ant-v2 (Figure \ref{fig:mujoco-tasks}(c)) for different values of $T_{\mbox{\scriptsize max}}$. The plots show that as $T_{\mbox{\scriptsize max}}$ increases, the deep RL policy presents a consistently improved behaviour. As RL policy is a function that maps the visited states to the selected actions, a closer deep RL policy to the optimal policy corresponds to a more predictable relationship between the training and the output trajectories. We argue that this feature of deep RL policies contributes to the higher level of vulnerability of the deep RL models that are trained with larger values of $T_{\mbox{\scriptsize max}}$. 

\begin{figure*}[h!]
\begin{center}
\includegraphics[width=0.7 \textwidth]{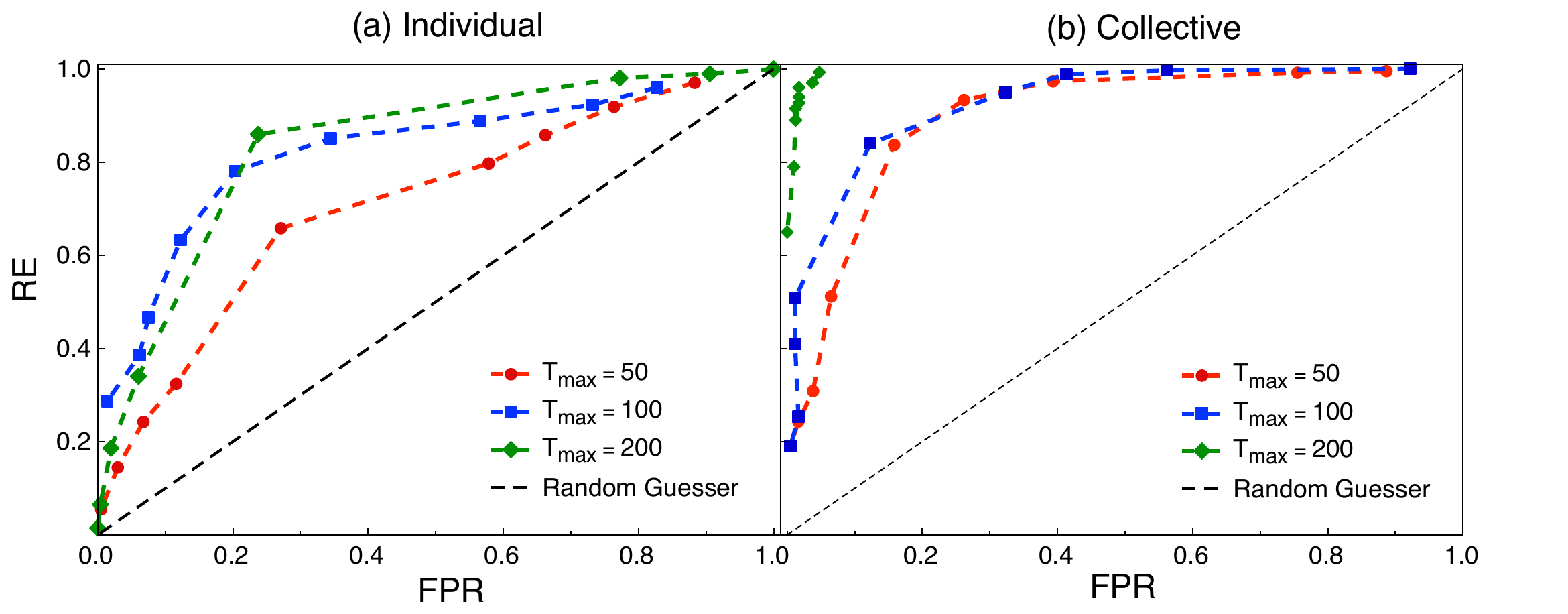} 
\end{center}
\caption{The receiver operator characteristic (ROC) curves of the MIA in HalfCheetah-v2 in the individual \textbf{(a)} and collective \textbf{(b)} modes for different values of $T_{\mbox{\scriptsize max}}$.} \label{fig:HC-ROC}
\end{figure*}
\begin{figure*}[b]
\begin{center}
\includegraphics[width=0.9\textwidth]{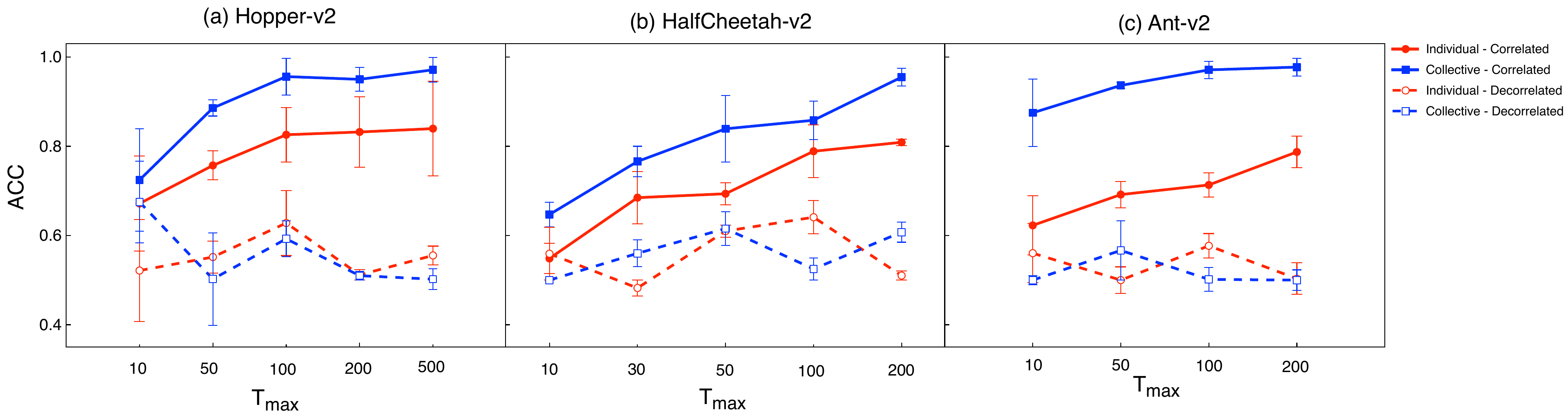} 
\end{center}
\caption{Comparison of the MIA accuracy between correlated and decorrelated settings for Hopper-v2 \textbf{(a)}, HalfCheetah-v2 \textbf{(b)}, and Ant-v2 \textbf{(c)}.} \label{fig:cor-decor}
\end{figure*}
As the attack classifiers output membership probabilities, we determine the predicted binary label for a range of acceptance thresholds $\theta = 0.1,~0.2,~\dots,~0.9$, and subsequently choose the threshold $\theta$, at which the classifier shows the highest performance. Figure \ref{fig:HC-ROC} depicts the sample ROC curves for HalfCheetah-v2 in individual (Figure \ref{fig:HC-ROC}(a)) and collective (Figure \ref{fig:HC-ROC}(b)) modes. The plots show that larger values of $T_{\mbox{\scriptsize max}}$ lead to better performance of the attack classifiers. The best result is obtained at $T_{\mbox{\scriptsize max}}=200$ in both individual and collective modes. We find that the acceptance threshold $\theta=0.5$ yields the highest performance throughout all of our experiments.

\subsection*{Temporal Correlation}
The results presented so far exhibit the performance of the MIAs against deep RL as a result of training the attack classifiers on the temporally correlated data collected from the training set and the output of the deep RL model. Considering that the training trajectories are decorrelated in the replay buffer as the first step after entering the deep RL trainer oracle, a question arises as to what role the temporal correlation in the data set plays in the vulnerability of deep RL models to MIAs. 

To answer this question, we have performed a set of experiments, where prior to the data augmentation phase, the temporal correlation between the deep RL training trajectories is broken. In particular, we decorrelate the training trajectories by shuffling their constituent tuples. We subsequently store the decorrelated transition tuples in an auxiliary buffer. In the next step, we generate trajectories of the desired length by sampling actions uniformly at random from the buffer. Finally, we pass the collection of decorrelated training trajectories together with output trajectories to the data augmentation mechanism and train the attack classifiers with the paired trajectories in the individual and collective modes. Figure \ref{fig:cor-decor} compares the accuracy of the MIA in the correlated mode with that in the decorrelated mode for the three tasks. The plots depict that the adversary's accuracy in inferring RL training members decreases significantly upon decorrelating the training trajectories. The results show that despite the inevitable input decorrelation imposed by the replay buffer mechanism in the training phase of off-policy deep RL models, the temporal correlation in the training trajectories is channelled to the model output data points. Thus, the attack classifiers trained on temporally correlated training data points exhibit higher accuracy than those trained on decorrelated trajectories.
\section{Conclusion} \label{sec:conclusion}

In this study, we design and evaluate the first membership inference attack (MIA) framework against off-policy deep RL in collective and individual membership inference modes by exploiting the inherent and structural temporal correlation present in deep RL data points. We demonstrate the performance of the proposed adversarial attack framework in complex high-dimensional locomotion tasks for different maximum trajectory lengths. Our proposed attack framework reveals the substantial vulnerability of a state-of-the-art off-policy deep RL model to the black-box MIAs. We show that it is significantly more vulnerable to MIA in the collective setting when compared to its vulnerability in individual MIAs. Moreover, our results demonstrate a consistent increase in the accuracy of the membership inference as a function of batch size in the collective mode. Furthermore, our experimental results reveal that the maximum trajectory length (in the episodic RL setting), which is set by the environment, plays a significant role in the vulnerability of the training data used in the deep RL model to the MIA. We show that a longer maximum trajectory length leads to an improved deep RL policy, thus a more defined relationship between the training and output trajectories, and consequently less private training data. Moreover, our results reveal the determinative role of temporal correlation in obtaining high MIA performance, which the attacker can utilize to design high-accuracy MIAs against deep RL. Despite the existence of replay memory as an intermediate data decorrelation mechanism at the heart of deep RL models, the trained policy still fully exploits the inherent correlation in learning feature representation, which poses a significant privacy concern in the deployment of trained RL policies at the industrial scale. Finally, the results from this study highlight serious privacy concerns in the widespread deployment of similar deep RL models, which demand more investigation of this matter to offer solutions in future studies. The tasks employed in the current study are under the umbrella of robotics simulation tasks that motivate the extension of experiments to real-world robot learning tasks. Moreover, dialogue systems such as Amazon Alexa, Apple Siri, and  Google Assistant are other interesting future platforms to apply RL-based MIAs on. In virtual dialogue systems, a data point is presented by a collection of interaction trajectories between the chatbot and the end user. A chatbot in this setting is the trained RL policy, and the user interactions with the bot form the training trajectories. In such settings, the collective mode is the natural inference setting since a collection of user interactions with the bot represents individual identity in the training set. In other words, the user's presence in the training set can be inferred by the adversary if and only if the attacker correctly infers a batch of trajectories representing the individual in the training set. Another extension to this line of research is to investigate MIAs against Deep RL models in a white-box setting, where the internal structure of the target policy is also known to the adversary.

\section*{Acknowledgements}
The authors would like to thank Hamidreza Ghafghazi and Spencer Main for their valuable contribution to the design and development of the preliminary version of the codebase. Computing resources were provided by Compute Canada, Calcul Qu\'ebec, and VIP Lab at the University of Waterloo throughout the project, which the authors appreciate. Funding was provided by the Natural Sciences and Engineering Research Council of Canada (NSERC).

\clearpage
\bibliographystyle{IEEEtran}
\bibliography{mybib}
\clearpage

\onecolumn
  \begin{center}
    {\bf APPENDIX}
  \end{center}
%\addcontentsline{toc}{chapter}{Appendices}% Print Appendix in ToC
\renewcommand{\thesection}{\Alph{section}}% Adjust section printing (from here onward)
\setcounter{section}{0}% Reset numbering for sections
%\appendix
%\appendixname{\begin{center}\large\textbf{Appendix}\end{center}}

In Section \ref{sec:arch}, we provide additional information regarding the attack network architectures we use to design attack classifier in the \emph{individual} and \emph{collective} modes. In Section \ref{sec:tbl}, we present the detailed results on the performance of the two attack classifiers in three standard continuous control locomotion MuJoCo tasks Hopper-v2 (Table \ref{tbl:Hopper_table}), HalfCheetah-v2 (Table \ref{tbl:HC_table}), and Ant-v2 (Table \ref{tbl:Ant_table}).

\section{Network Architectures}\label{sec:arch}

\textbf{Individual-Mode Attack Classifier Architecture- } In the individual  mode, we use Temporal Convolutional Network (TCN) architecture \cite{bai2018empirical}, which takes temporally correlated data as input and uses a hierarchy of dilated (causal) convolutional layers to capture the inherent temporal correlation in the input data. In this study, since the input data to the classifier in the individual mode is a pair of temporally correlated tensors (\emph{i.e.} $\R^{2d^{\actions} \times T }$), the size of \emph{input channel} in TCN architecture represents twice the dimension of the action space. In the design of the TCN architecture, we use a two-layer network and set the number of hidden layers to 600. We set the kernel size to 3 and use dropout with a threshold of 0.45. For network training, we employ Adam optimizer \cite{adam} with the initial learning rate set at 0.0003 and gradient clipping at 0.35 to avoid exploding gradients. We change the initial learning rate during the training phase using the learning rate scheduler to avoid local minimum. In this regard, we use a learning rate scheduler at 100 and 200 epochs with gamma rate 0.1 to decay the learning rate with time. The batch size is set to 16 during training, and the network is trained for 300 epochs.\\

% In addition, TCN makes the representation of higher-dimension input easier by employing an input channel. 
% We notice that the lower the batch size, the higher the attack's success rate.

\noindent\textbf{Collective-Mode Attack Classifier Architecture- } In the collective mode, we use deep Residual Networks (ResNets) as the choice of attack classifier. In particular, we employ ResNet-18 architecture, where in addition to an input channel, there is a width dimension, which we use to input a collection of trajectories as a batch. Our input is in the form of a three-dimensional tensor (\emph{e.g.} $\R^{2d^{\actions} \times T \times m }$), where $m$ denotes the number of trajectories in each batch, similar to the data structure used in image classification problems \cite{he2016deep, simonyan2014very, huang2017densely}. Moreover, we use Adam optimizer with learning rate 0.0008 and weight decay 1.0, and we set the clipping size to 0.35. Finally, we use batch size 256 and train the network for 300 epochs.
\section{Attack Performance}\label{sec:tbl}
In Table \ref{tbl:Hopper_table}, we show the results obtained for the attack performance on the deep RL algorithm in the three MuJoCo environments in terms of five performance metrics \emph{Accuracy}, \emph{Precision}, \emph{recall}, \emph{F1 score}, and \emph{Matthews correlation coefficient}.
\begin{table}[h!]
    \caption{The performance of the attack classifiers in Hopper-v2 for different maximum trajectory lengths $T_{\mbox{\scriptsize max}}$ in terms of accuracy (ACC), precision (PR), recall (RE), F1 score (F1), and Matthews correlation coefficient (MCC). The values in parentheses show the results for the collective attack mode.}
    \label{tbl:Hopper_table}
    \centering
    \includegraphics[width=0.65\linewidth]{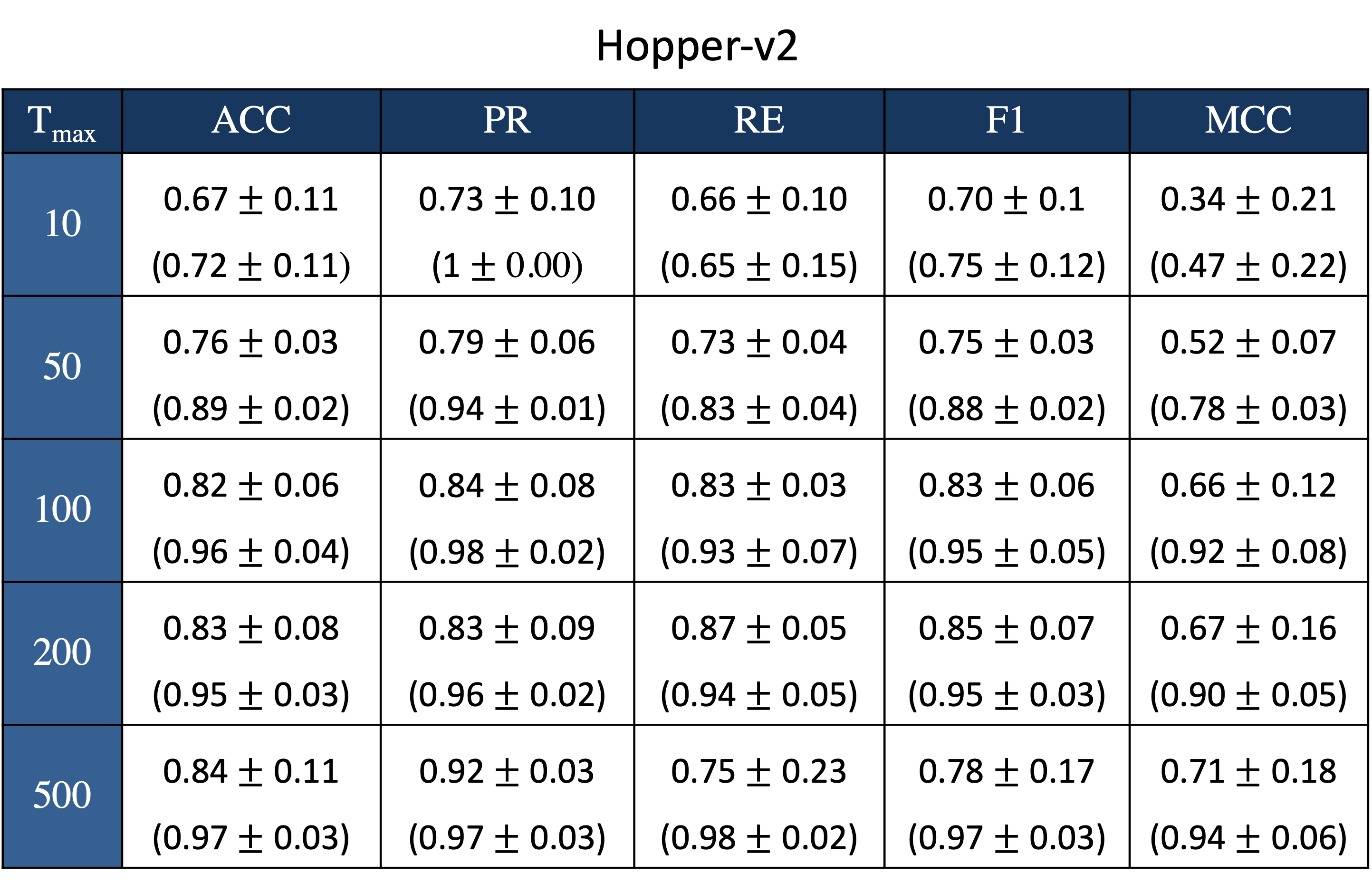}
  \end{table}

  \begin{table}[h!]
    \caption{The MIA performance results for individual and collective modes in HalfCheetah-v2 for different maximum trajectory lengths $T_{\mbox{\scriptsize max}}$ in terms of accuracy (ACC), precision (PR), recall (RE), F1 score (F1), and Matthews correlation coefficient (MCC). The values in parentheses present the results for the collective attack mode.}
    \label{tbl:HC_table}
    \centering
    \includegraphics[width=0.65\linewidth]{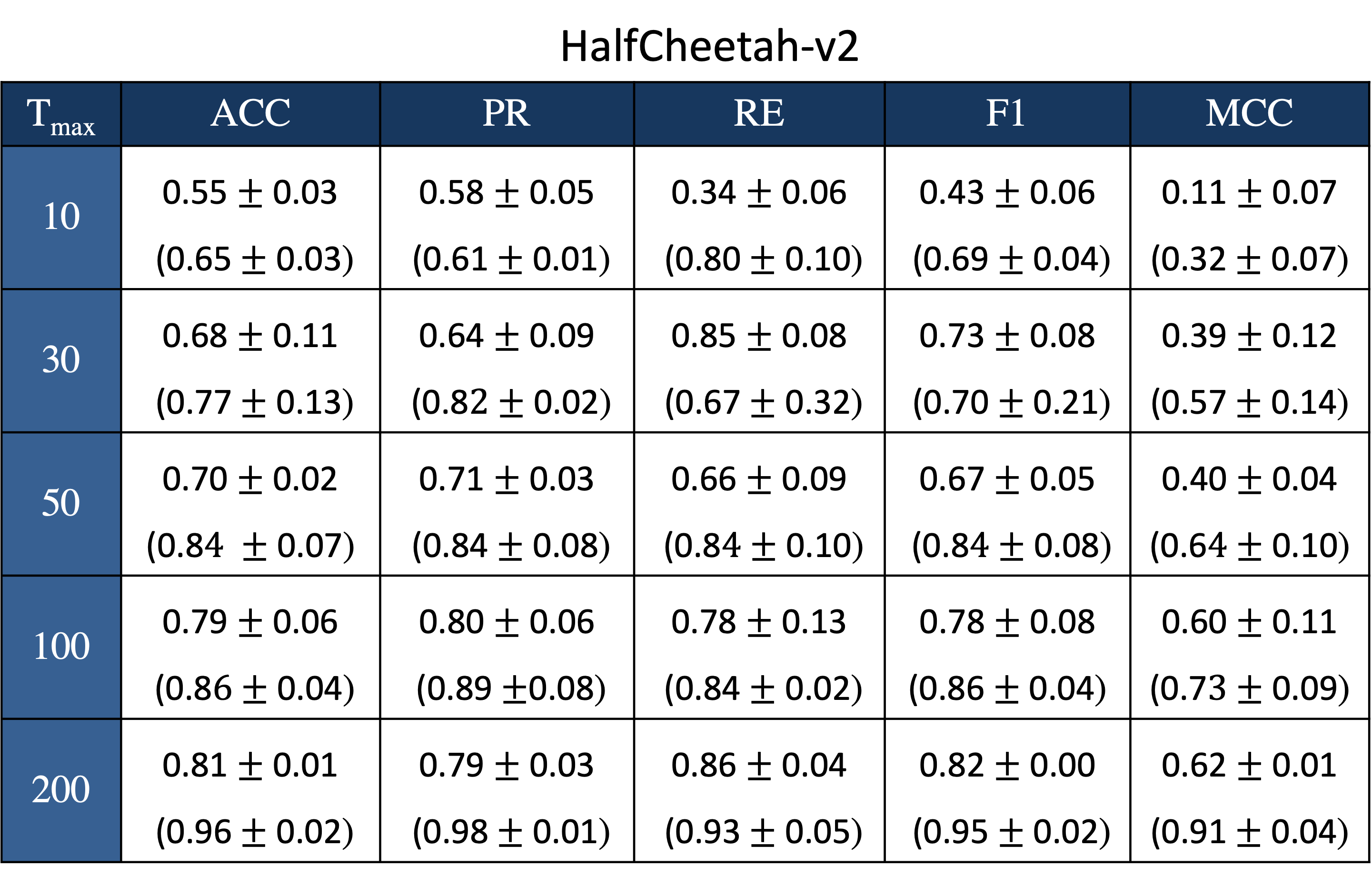}
  \end{table}

  \begin{table}[h!]
    \caption{The performance of the attack classifiers in Ant-v2 for different maximum trajectory lengths $T_{\mbox{\scriptsize max}}$ in terms of accuracy (ACC), precision (PR), recall (RE), F1 score (F1), and Matthews correlation coefficient (MCC). The values in parentheses show the results for the collective attack mode.}
    \label{tbl:Ant_table}
    \centering
    \includegraphics[width=0.65\linewidth]{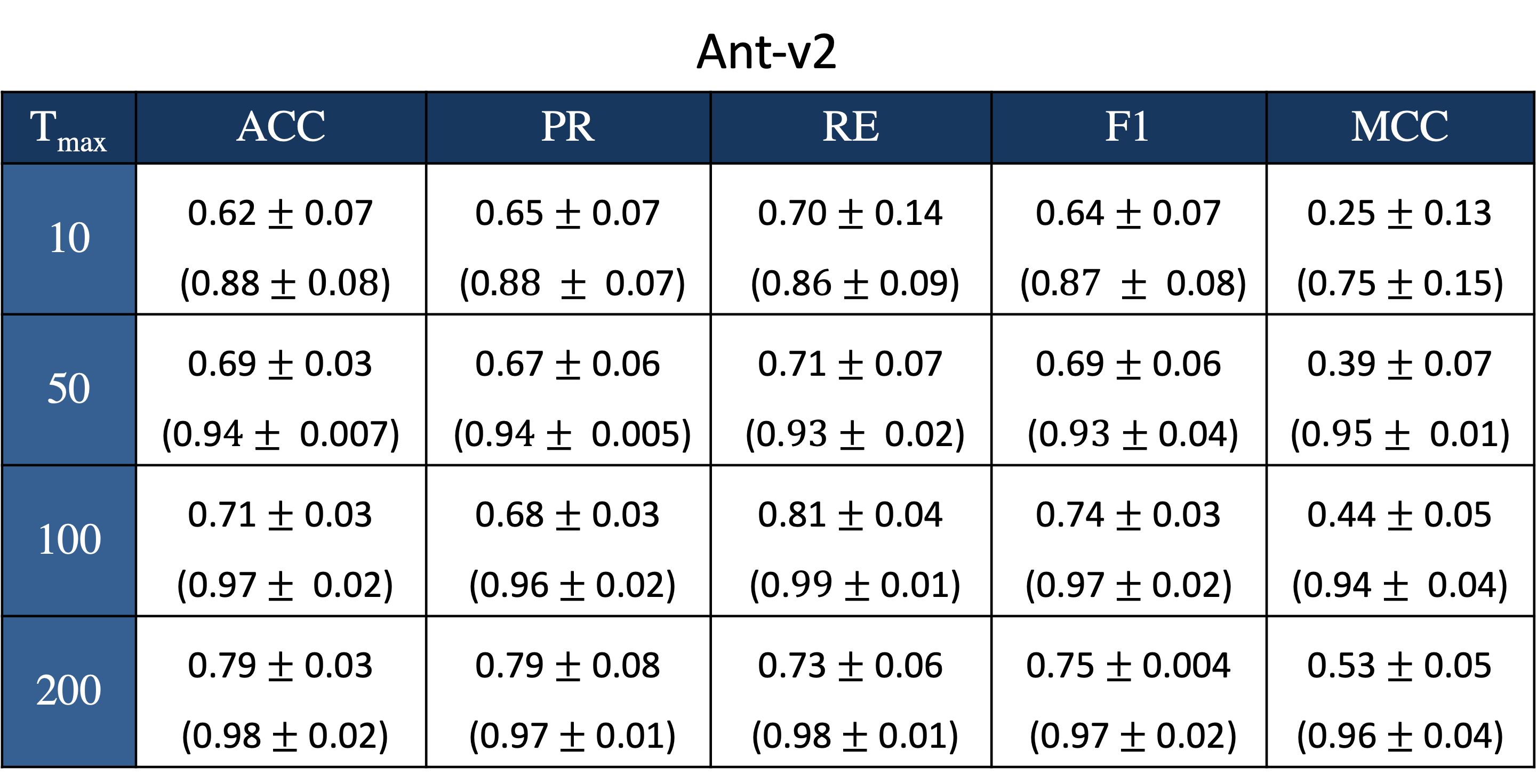}
  \end{table}

\end{document}